\documentclass[runningheads]{llncs}
\usepackage{graphicx}

\usepackage{multirow}
\usepackage{caption}
\usepackage{subcaption}

\begin{document}
\title{Learning from Auxiliary Sources in Argumentative Revision Classification\thanks{Supported by the National Science Foundation under Grant \#173572.}}
%
%
\author{Tazin Afrin
\and
Diane Litman
}
\authorrunning{Afrin et al.}
%
\institute{University of Pittsburgh, Pittsburgh PA 15260, USA \\
\email{\{taa74,dlitman\}@pitt.edu}}
\maketitle              
\begin{abstract}
We develop models to classify desirable reasoning revisions in  argumentative writing. We explore two approaches -- multi-task learning and transfer learning -- to take advantage of  auxiliary sources of revision data for similar tasks. Results of intrinsic and extrinsic evaluations show that both approaches can indeed improve classifier performance over baselines. While multi-task learning shows that training on different sources of data 
at the same time may improve performance, transfer-learning better represents the relationship between the data.

\keywords{Writing \and Revision \and Natural Language Processing.}
\end{abstract}
\section{Introduction}
Our research focuses on the {\it automatic classification of desirable 
revisions of reasoning}\footnote{Such revisions of text {\it content} 
are considered more useful in revising~\cite{zhang2017hh}.} in argumentative writing. By reasoning, we refer to how  evidence is explained and linked  to an overall argument. Desirable revisions (e.g., reasoning supporting the evidence) are those that have hypothesized utility in improving an essay, 
while undesirable revisions do not have such hypothesized utility~\cite{afrin2023eacl}. Identifying desirable revisions should  be helpful for improving intelligent feedback generation in automated writing evaluation (AWE) systems 
\cite{roscoe2015automated}.
In this study, we focus on improving our model learning by taking advantage of auxiliary data sources of revisions. For example, we would like to see how college-level essay data might be beneficial for elementary-school essays. 
We train two types of models -- a {\it multi-task learning} (MTL) model, and a {\it transfer learning} (TL) model. In our MTL experiment, we allow information sharing during training using different source data. In our TL experiment, we fine-tune a model pre-trained on  source data to see  which type of source data might improve  performance on the target data. 
Our results show that both MTL and TL are beneficial for the datasets written by more novice writers. However, for more expert writers (e.g., college students), it is difficult to further improve classifier  performance.

%
%
%

Prior Natural Language Processing (NLP) research 
in academic writing has focused on classifying argumentative revision purposes~\cite{zhang2017hh} 
and understanding revision patterns~\cite{roscoe2015automated}. 
While some have classified 
revisions in terms of {\it quality}~\cite{tan2014l},
~\cite{afrin2020RER} were the first 
to consider a revision's
utility in improving an essay in alignment with previously received  feedback. They also investigated state-of-the-art models to identify desirable revisions of evidence and reasoning on three different corpora~\cite{afrin2023eacl}. 
We extend their revision annotation framework on one additional corpus, as well as {\it leverage the corpora as auxiliary sources for MTL and TL}.
MTL in NLP is widely used to learn from a limited data source (e.g., 
~\cite{schulz2018-multi-argmining} used the same primary task of argument classification for multiple low-resource datasets  framed as MTL). 
Following~\cite{schulz2018-multi-argmining}, we 
explore MTL to classify desirable reasoning revisions for multiple datasets. To the best of our knowledge this is the {\it first exploration of  MTL for revision classification}. 
Transfer learning in NLP is used to reduce the need for labeled data by learning representations from other models~\cite{chakrabarty2019-ampersand,ghosh2020YoungArgument}. 
Unlike previous works, we first train our model using \textit{source revision data}, then fine-tune the model for a \textit{target revision data}.

\section{Data and Annotations}
\label{section: data and resources}

\begin{table}[t]
\caption{Comparison of essay corpora used in this study.} 
\label{table:datasets}
\centering

\begin{tabular}{|l|c|c|c|c|c|c|}
\hline											
\multirow{2}{*}{Corpus}	&	\multirow{2}{*}{\#Students}	& Grade	& Feedback &		Essay Drafts		&   Essay Score 	& Improvement \\
	&		&             Level     & Source        	&   Used	&  Range &  Score Range\\ \hline
E	& 143   &	 $5^{th}$ \& $6^{th}$   & AWE	&   1 and 2	&   [1, 4]   &  [0, 3]\\ \hline
H1	&	47  & 	  $12^{th}$           &  peer	&   1 and 2	&   [0, 5]  &  [-2, +3] \\ \hline
H2	&	63  & 	  $12^{th}$          &   peer	&   1 and 2	&   [17, 44]  &  [-14, +12] \\ \hline

C	    &	60	&       college      & AWE      	&   2 and 3	&   [15, 33]    &  -1, +1\\ \hline
\end{tabular}

\end{table}

Our data consists of reasoning revisions from four corpora of paired drafts of argumentative essays used in previous revision classification tasks~\cite{zhang2017hh,afrin2023eacl}.  All essays were written by students in response to a prompt, revised in response to feedback, and graded with respect to a rubric. 
A corpus comparison is shown in Table~\ref{table:datasets}. Their diversity along multiple dimensions makes it challenging to train one model for all. However, since our target is to classify  revisions following one annotation framework, it is compelling to investigate how these datasets might be related. In the {\it elementary} (E) school corpus, students wrote Draft1 about a project in Kenya, then received feedback from an automated writing evaluation (AWE) system. All essay pairs were later graded on a scale from 0 to 3 to indicate improvement from Draft1 to Draft2 in line with the feedback~\cite{afrin2020RER}. Two corpora contain essays written by  {\it high-school} students and revised in response to peer feedback -- H1 and H2. Drafts 1 and 2 of each high-school corpus were  graded using separate rubrics by expert graders. We create an improvement score for each essay pair, calculated  as the difference of the holistic score between drafts. 
The {\it college} (C) essays unlike the other  corpora involving proprietary data, were downloaded from the web~\cite{zhang2017hh}. 
Students received general feedback  after Draft1,
revised to create Draft2, then revised again to create Draft3 after receiving essay-specific feedback from an AWE system. We create a binary improvement score, calculated as 1 if Draft3 improved compared to Draft2, -1 otherwise.

For all essays in each corpus, sentences from the two drafts were aligned manually based on semantic similarity.
Aligned sentences represent one of four operations between drafts -- no change, modification, sentence  deleted from Draft1, sentence added to Draft2. Each pair of changed aligned sentences was then extracted as a \textit{revision} and annotated for its {\it purpose} (e.g., revise reasoning), using the scheme introduced for the college corpus~\cite{zhang2017hh}. From among the full set of annotations, we only use reasoning revisions for the current study because they are the most frequent across the four corpora. 
The reasoning revisions were then annotated for its desirability~\cite{afrin2023eacl}. We leverage the annotated E, H1, and C data from the previous study~\cite{afrin2023eacl} and extend the annotation of a new data, H2. 

\begin{table}[t]
\caption{Average number of revisions over 10-fold cross-validation is shown before and after data augmentation (D = Desirable, U = Undesirable).}
\label{table: data distribution}
\centering
\begin{tabular}{|l|c|c|c|c|c|c|c|c|c|c|}
\hline
	&  &	\multicolumn{3}{c|}{Before Augmentation}	&	\multicolumn{3}{c|}{Augmented for MTL}	&	\multicolumn{3}{c|}{Augmented for TL}	\\	\hline
Corpus	& N  & 	Total	&	D	&	U	&	Total	&	D	&	U &	Total	&	D	&	U	\\	\hline
E	& 143 &	389	&	186	&	203	&	5120	&	2376	&	2744	&	7725	&	3881	&	3844\\	
H1	& 47 &	387	&	202	&	185	&	5120	&	2750	&	2370	&	5780	&	2963	&	2817\\	
H2	& 63 &	329	&	169	&	160	&	5120	&	2770	&	2350	&	10986	&	5997	&	4989\\	
C	& 60 &	207	&	114	&	93	&	5120	&	2894	&	2226	&	5515	&	3186	&	2329\\	\hline
\end{tabular}
\end{table}

Deep learning requires more than our limited amount of data for training. We use the synonym replacement data augmentation strategy  from our prior work~\cite{afrin2023eacl}  to generate more training examples. Since MTL is trained batch by batch for each data in a round robin fashion, we selected a fixed number of instances from each dataset to stay consistent. TL used all available data. Table~\ref{table: data distribution} shows the number of desirable and undesirable revisions for each corpus.

\section{Models}
\textbf{Single-task Learning Model (STL).}
Our STL model is a neural network model used in previous desirable revision classification task~\cite{afrin2023eacl}. The input to the model is the revision sentence pair. The model uses the pre-trained BERT (`bert-base-uncased') embedding with a BiLSTM and a Dense layer. The output is a sigmoid activation function for the binary classification task. Classifying desirable reasoning in each corpus is considered an individual task due to the difference in corpora summarized in Table \ref{table:datasets}. Following previous work, we also select the learning rate  $1e^{-3}$ and batch size 16, and apply the same to all data.

\textbf{Multi-task Learning Model (MTL).} 
The individual tasks in STL are combined in MTL with a \textit{shared BiLSTM layer}. After encoding the revision using the off-the-shelf BERT encoder, we send this to the BiLSTM layer. The BiLSTM layer learns shared information between different tasks. Each task has an individual Dense layer and a Sigmoid output layer to learn task-specific information. 
During training, we use the same settings as in STL. In MTL, we train the model in the sequence of C, H1, H2, and E data in a round robin fashion for each batch. 
This sequence is repeated for all the batches for 10 epochs. Since our batch size is very small, we believe the training will not be affected substantially by the order of selecting the data. We apply 10-fold cross-validation. During testing, we use the respective task for the respective data. 

\textbf{Union Model.} 
Unlike STL (where we train a separate model for each of the four corpora/tasks), for the Union Model we train only one STL model. We use the union of all task data as input. The training is performed following the MTL (e.g., batch by batch) training process. 
Compared to STL, the Union model will help us understand if using extra data as a source of information is beneficial. Comparison of the Union and MTL models will help us  validate that any MTL improvement is not just due  to more training data. 

\textbf{Transfer Learning Models (TL).} 
In transfer learning, we learn from a source dataset and apply it to a target dataset to understand the relationship between our data, which may not be obvious from MTL. The source and target data are taken from all possible combinations of  our datasets (Table~\ref{table:datasets}). 
TL also adopts the STL model to first train with the source data, then fine-tune with the target data 
using all the augmented data available (shown in Table~\ref{table: data distribution}).

\section{Results}
\label{sec: results}
In our \textit{intrinsic} evaluation, classification performance was compared to baseline models in terms of average unweighted F1-score over 10-folds of cross-validation. We compare MTL with two baselines (STL and Union), while TL was compared against one baseline (STL trained on target data). 
\textit{Extrinsic} evaluation checked how often desirable and undesirable revisions (gold annotations) are related to improvement score using Pearson correlation. We then replicate the process for the predicted revisions to see if they are also  
correlated in the same way.

\textbf{MTL evaluation.} Intrinsic evaluation in Table~\ref{table: intrinsic mdtl} shows that in-general MTL has higher average f1-scores. However, MTL and baseline  results are close with  no significant difference. 
\begin{table}[!t]
\centering
\caption[Evaluation results.]{MTL and TL evaluation.  Best results are bolded. $\uparrow$ indicates TL improved over STL. Extrinsic evaluation shows Desirable results only. Significant predicted correlations consistent with using gold labels are bolded.   *  p$<.05$.}
\label{table: mtl and tl evaluation}
\begin{subtable}{.45\linewidth}
\caption[Intrinsic Evaluation: using multi-task-learning.]{ MTL Intrinsic Evaluation}
\label{table: intrinsic mdtl}
\centering
\begin{tabular}{|l|c|c|c|}
\hline
Corpus	&		STL & Union & MTL	\\	\hline
E	&	0.597 & 0.583 & 	\textbf{0.607}  \\ \hline
H1	& \textbf{0.649} & 0.631 & 0.627	\\ \hline
H2	&	0.633	&	0.622	&	\textbf{0.658}	\\	\hline
C	&	0.613	&	0.539	&	\textbf{0.619}	\\	\hline

\end{tabular}
\end{subtable} %
\begin{subtable}{.45\linewidth}
\caption[Extrinsic Evaluation: using multi-task-learning.]{MTL Extrinsic Evaluation}
\label{table: predicted revision correlation mdtl}
\centering
\begin{tabular}{|l|l|l|l|l|}
\hline									
	&		Gold	&	STL     &   Union	&	MTL	\\	\hline	\hline
	
E		&	\textbf{0.450*}	&	\textbf{0.339*}	&	\textbf{0.347*}	&	\textbf{0.317*}	\\		
 \hline
H1		&	\textbf{0.351*}	&	0.249	&	0.266	&	0.222	\\		
 \hline
H2		&	\textbf{0.301*}	&	\textbf{0.274*}	&	\textbf{0.300*}	&	0.232	\\		
 \hline
C		&	0.029	&	0.039	&	-0.057	&	0.003	\\		
 \hline
\end{tabular}
\end{subtable}
\\
\begin{subtable}{.45\linewidth}
\caption[Intrinsic Evaluation TL]{TL Intrinsic Evaluation.}
\label{table: intrinsic tl}
\centering
\begin{tabular}{|c|l|c|c|c|c|c|}
\hline	
&   &	\multicolumn{4}{c|}{Target}		\\	\cline{3-6}
	& &	E	&	H1	&	H2	&	C	\\	\cline{3-6} \hline
& STL	&	0.597	&	0.649	&	0.633	&	0.613	\\	\hline \hline
\multirow{4}{*}{\rotatebox[origin=c]{90}{Source}} & E	&		&	0.661$\uparrow$	&	\bf{0.652}$\uparrow$	&	0.606	\\	\cline{2-6}
& H1	&	0.607$\uparrow$	&		&	0.636$\uparrow$	&	\bf{0.644$\uparrow$}	\\	\cline{2-6}
& H2	&	0.606$\uparrow$	&	{\bf 0.678}$\uparrow$	&		&	\bf{0.644$\uparrow$}	\\	\cline{2-6}
& C	&	\bf{0.641$\uparrow$}	&	0.638	&	0.598	&		\\	\hline
\end{tabular}

\end{subtable}
\begin{subtable}{.45\linewidth}
\caption[Extrinsic Evaluation TL]{TL Extrinsic Evaluation.}
\label{table: extrinsic tl}
\centering
\begin{tabular}{|c|l|c|c|c|c|}
\hline	
	&		&	\multicolumn{4}{c|}{Target}		\\	\cline{3-6}
	&		&	E		&	H1		&	H2	&	C		\\	
 \hline \hline
	&	Gold	&	\textbf{0.450*}	&	\textbf{0.351*}		&	\textbf{0.301*}	&		0.029		\\	\cline{2-6}
	&	STL 	&	\textbf{0.339*}	&	0.249		&	\textbf{0.274*}	&	0.039		\\	\hline \hline
\multirow{4}{*}{\rotatebox[origin=c]{90}{Source}} 	&	E	&		&	0.262	&	\textbf{0.262*}	&	0.033		\\	\cline{2-6}
	&	H1	&	\textbf{0.337*}		&	&	\textbf{0.308*}	&	0.008		\\	\cline{2-6}
	&	H2	&	\textbf{0.360*}		&	\textbf{0.376*}		&		&	-0.060		\\	\cline{2-6}
	&	C	&	\textbf{0.350*}		&	\textbf{0.292*}	&	\textbf{0.250*}	&		\\	\hline
\end{tabular}
\end{subtable}
\end{table}
Further investigation showed MTL outperformed baselines in identifying undesirable revisions. MTL also showed improvement over Union baseline,  indicating that MTL's success over STL is not just due to more training data. Union performed worse than STL, emphasizing the importance of data usage.
In extrinsic evaluation,  MTL showed significant positive correlations for predicted desirable reasoning for the E data (Table~\ref{table: predicted revision correlation mdtl}), which is consistent with the Gold correlation. 
In other cases, either MTL is not consistent with Gold, or the correlation is not to be significant. In contrast, both STL and Union often showed significant correlation to essay improvement.
Our results suggest treating our datasets as individual tasks to better relate to student writing improvement (extrinsic evaluation). However, we found sharing features (via MTL) useful for identifying desirable reasoning (intrinsic evaluation).

\textbf{TL evaluation.}
Elementary-school students can be considered as the least experienced writers in our datasets considering the age group. Hence the result in Table~\ref{table: intrinsic tl} may indicate that model for elementary-school students needed to learn the structure from better-written essays. 
Unlike how the H2 data as a source helped the H1 data as the target, the reverse is not entirely true. 
Moreover, transfer from C decreased performance for H2. 
Finally, for C data, transfer learning the weights from both high-school datasets helped improve performance over the baseline (trained only on the target data). 
Although E as the target domain improved most when learning from the C data, the reverse is not true. College-level students were comparatively experienced writers in our corpora, so inexperienced student writing may have not helped. 

Table~\ref{table: extrinsic tl} shows that when our target is the E data, TL results are consistent with Gold annotation results for desirable reasoning. Undesirable revisions are not significant. 
When H1 is the target, 
H2 yields the highest correlation, which might be because it is also high-school data. 
C data also showed significant correlation in this case. 
While all models are consistent with Gold annotation for H2 as the target, H1 showed the highest correlation of desirable revision to essay improvement score. This is surprising because H1 did not improve in intrinsic evaluation. 
Finally, C as target did not see any significant correlations.

Overall, transfer learning shows that the availability of more data or more information is not enough. Rather, which data is used to pre-train or how it is being used to train the model (e.g., MTL) is also important. Extrinsic evaluation also supports the fact that more data does not mean improvement. For example, transfer from other high-school data yield stronger results for H1 or H2 data compared to transfer from E or C data.
Overall, our results from the transfer learning experiments show that for each target data there were one or more source corpora that improved the classifier performance. 

\section{Discussion of Limitations and Conclusion}

Our corpora were originally annotated using a detailed revision scheme~\cite{afrin2020RER}, then used a simplified scheme~\cite{afrin2023eacl} to create binary revision classification task. 
In a real-world scenario, an end-to-end AWE system deploying our model 
would have errors propagated from alignment and revision purpose classification and perform lower than the presented model. Moreover, we need to  examine our additional but less frequent revision purposes too. 
We also plan to explore other data augmentation techniques to experiment with more complex models. 
Although we do not  have demographic information, the students in the college corpus include both native English and non-native speakers \cite{zhang2017hh}.
Another limitation is that the MTL model training process is  slow. 
The current methods also require GPU resources. 
Moreover, we only investigated one sequence of training the MTL. 

%
%
%
%

We explored the utility of predicting the desirability of reasoning revisions using auxiliary sources of student essay data using multi-task learning and transfer learing. 
Both 
experiments indicate that there is common information between datasets that may help improve 
classifier performance. 
Specifically, the results of our intrinsic and extrinsic evaluations  show that while desirable revision classification using auxiliary sources of training data can improve performance, 
the data from 
different argumentative writing tasks needs to be utilized wisely.

%
%
%
\bibliographystyle{splncs04}
\bibliography{references}
\end{document}